\documentclass[sigconf,screen]{acmart}

\AtBeginDocument{%
  \providecommand\BibTeX{{%
    \normalfont B\kern-0.5em{\scshape i\kern-0.25em b}\kern-0.8em\TeX}}}

\setcopyright{acmcopyright}
\copyrightyear{2018}
\acmYear{2018}
\acmDOI{10.1145/1122445.1122456}

\acmConference[Woodstock '18]{Woodstock '18: ACM Symposium on Neural
  Gaze Detection}{June 03--05, 2018}{Woodstock, NY}
\acmBooktitle{Woodstock '18: ACM Symposium on Neural Gaze Detection,
  June 03--05, 2018, Woodstock, NY}
\acmPrice{15.00}
\acmISBN{978-1-4503-XXXX-X/18/06}

\begin{document}

\title{Gophormer: Ego-Graph Transformer for Node Classification}

\author{Jianan Zhao}
\authornotemark[1]
\affiliation{%
  \institution{University of Notre Dame}
  \city{Notre Dame}
  \country{USA}}
\email{jzhao8@nd.edu}

\author{Chaozhuo Li}
\affiliation{%
  \institution{Microsoft Research}
  \city{Beijing}
  \country{China}
}
\email{cli@microsoft.com}

\author{Qianlong Wen}
\authornote{Both authors contributed equally to this research.}
\affiliation{%
  \institution{University of Notre Dame}
  \city{Notre Dame}
  \country{USA}}
\email{qwen@nd.edu}

\author{Yiqi Wang}
\affiliation{%
  \institution{Michigan State University}
  \city{East Lansing}
  \country{USA}}
\email{wangy206@msu.edu}

\author{Yuming Liu}
\affiliation{%
  \institution{Microsoft}
  \city{Beijing}
  \country{China}
}
\email{yumliu@microsoft.com}

\author{Hao Sun}
\affiliation{%
  \institution{Microsoft}
  \city{Beijing}
  \country{China}
}
\email{hasun@microsoft.com}

\author{Xing Xie}
\affiliation{%
  \institution{Microsoft Research}
  \city{Beijing}
  \country{China}
}
\email{xingx@microsoft.com}

\author{Yanfang Ye}
\authornote{Corresponding author.}
\affiliation{%
  \institution{University of Notre Dame}
  \city{Notre Dame}
  \country{USA}
}
\email{yye7@nd.edu}

\renewcommand{\shortauthors}{Anonymous Author(s).}
\newcommand{\czl}[1]{{\color{red}#1}}
\newcommand{\zja}[1]{{\color{blue}#1}}
\newcommand{\yq}[1]{{\color{green}#1}}
\newcommand{\qlw}[1]{{\color{blue}#1}}

\begin{abstract}
 Transformers have achieved remarkable performance in a myriad of fields including natural language processing and computer vision. However, when it comes to the graph mining area, where graph neural network (GNN) has been the dominant paradigm, transformers haven't achieved competitive performance, especially on the node classification task. Existing graph transformer models typically adopt fully-connected attention mechanism on the whole input graph and thus suffer from severe scalability issues and are intractable to train in data insufficient cases. To alleviate these issues, we propose a novel Gophormer model which applies transformers on ego-graphs instead of full-graphs. Specifically, Node2Seq module is proposed to sample ego-graphs as the input of transformers, which alleviates the challenge of scalability and serves as an effective data augmentation technique to boost model performance. Moreover, different from the feature-based attention strategy in vanilla transformers, we propose a proximity-enhanced attention mechanism to capture the fine-grained structural bias. In order to handle the uncertainty introduced by the ego-graph sampling, we further propose a consistency regularization and a multi-sample inference strategy for stabilized training and testing, respectively. Extensive experiments on six benchmark datasets are conducted to demonstrate the superiority of Gophormer over existing graph transformers and popular GNNs, revealing the promising future of graph transformers.
\end{abstract}




\maketitle

\section{Introduction}
\label{sec: Intro}
In recent years, graph neural networks (GNNs), have shown excellency in graph mining and are widely applied to a variety of applications such as node classification~\cite{GCN,GAT,GraphSAGE}, graph classification~\cite{GIN,GraphSAINT}, and recommendation~\cite{PinSAGE,GraphRec,NGCF}. Most GNN methods follow a message-passing scheme where the node representation is learned by aggregating and transforming its neighborhood embeddings~\cite{GCN,GNNSurvey}. 
Despite the prevailing adoption of the message-passing scheme, there is a growing recognition of the inherent limitations of this paradigm. On one hand, due to the repeated aggregation of local information, the message-passing scheme suffers from the over-smoothing issue, i.e. representations of different nodes become indistinguishable when stacking too many feature propagation layers~\cite{Oversmoothing_p1}. On the other hand, due to the exponential blow-up in computation paths as the model depth increases \cite{Bottleneck}, GNNs also experience difficulties in capturing long-range interactions (the over-squashing problem).

Meanwhile, in the fields of natural language processing and computer vision, the transformer architecture has shown dominant performance \cite{Transformer,TransformerXL,VIT,SwinTFM} thanks to its powerful feature learning capacity and low inductive bias. In light of its amazing performance and the limitations of GNNs, attempts have been made to introduce transformer into graph learning to replace the message-passing scheme. The majority of these methods \cite{GT,SAN,Graphormer} apply transformers on the entire graph, treating all nodes as fully-connected, and enhance the vanilla feature-based attention mechanism with structural encoding and topology-enhanced attention mechanism. For example, SAN \cite{SAN} learns Laplacian positional encodings and uses different attention mechanism for connected and unconnected nodes. Graphormer~\cite{Graphormer} achieves state-of-the-art graph classification performance via transformers with centrality, spatial, and edge encodings. However, there are several limitations of these methods. First, the full-graph attention schema suffers from the quadratic dependency (mainly in terms of memory) on the graph size due to the full attention mechanism, leading to the poor scalability especially on large graphs. 
Moreover, the full-attention mechanism views the entire input graph as a fully-connected graph and fuses information from all the nodes, which potentially introduces noise from numerous irrelevant long-distance neighbors.
Last but not least, unlike GNNs with few learnable parameters, the transformers contain much more parameters, which are intractable to be fully trained with scarce annotations and are also more vulnerable to overfitting.

In light of the potential noise from the long-distance neighbors and the challenge of scalability, we propose to use the ego-graph instead of the entire graph as the transformer input. Specifically, we propose a novel module named Node2Seq to sample an ego-graph for each node as the input of the transformer. Then, node representations are learned based on the features within the sampled ego-graphs instead of node features of the entire graph.
This strategy apparently alleviates the aforementioned problems: First, the challenge of scalability is eased since the input size is reduced from $|V|$ (number of nodes in the whole graph) to $|V_{E}|$ (average number of nodes in ego-graphs) with $|V| \gg |V_{E}|$. In this way, transformers can be easily applied to large graphs. Second, the ego-graph preserves the localized contextual information within a pre-defined order, which consequently filters the high-order noise. 
Last but not least, given the data-hungry~\cite{datahungry, GPT3} (more data leads to better performance) characteristic of transformers, the sampling process can be viewed as an effective data augmentation operator, which greatly boosts the performance of graph transformers. 
Despite these benefits, several challenges still need to be addressed when using ego-graph-based transformer. To start with, due to the fully-connected nature of the attention mechanism, the structural information is lost in the construction of sequential input. Therefore, we need to design an effective attention mechanism to incorporate the vital structural information. Moreover, though more scalable, the constructed ego-graph of each node aggregates information solely from its local neighbors and neglects the high-order information of graphs. Hence we need to design another strategy to incorporate the global information as complementary. Last but not least, the sampled ego-graphs of a center node is essentially a subset of this node's full-neighbor ego-graph, which may lost important information and renders potentially unstable performance. 

To address the aforementioned challenges, in this paper we propose a novel model dubbed E\underline{go}-gra\underline{ph} Transf\underline{ormer} (Gophormer) to learn desirable node representations. We demonstrate that using sampled ego-graphs instead of conventional full-graph greatly boosts the transformer's performance. Specifically, Gophormer constructs ego-graphs integrated with global nodes by Node2Seq to capture both local and global structural information. 
After that, proximity-enhanced attention mechanism is proposed to incorporate both feature and structural proximity to learn node embeddings. 
To alleviate the uncertainty caused by sampling module, the ego-graph consistency regularization is applied in the training phase to regularize different samples, and the multi-sample inference strategy is proposed to stabilize predictions. 
It is noteworthy to highlight our contributions as follows:
\begin{itemize}
    \item We propose to use Node2Seq that converts graph data to ego-graph-based sequential input for transformers. Node2Seq not only enables scalable training but also successfully boosts existing graph transformer frameworks on node classification tasks significantly. 
    \item We propose a novel model Gophormer. Gophormer utilizes Node2Seq to generate input sequential data and encodes it with proximity-enhanced transformer. To alleviate the impact of uncertainty, we design ego-graph consistency regularization at training time and propose to use multi-sample inference during testing time.
    \item Extensive experiments are conducted on six benchmark datasets. Gophormer achieves state-of-the-art performance, demonstrating the great potential of graph transformers.
\end{itemize}

\section{Background}
\label{sec: rw}
Given an input graph $G=(V,E)$ composed of a node set $V$ and an edge set $E$, its graph structure information can be represented as an adjacency matrix $\mathbf{A} \in\{0,1\}^{|V| \times|V|}$, where $|V|$ denotes the number of nodes in the graph. The element $\mathbf{A}[i, j]$ in the adjacency equals to 1 if there exists an edge between node $v_i$ and $v_j$, otherwise $\mathbf{A}[i, j]=0$. Each node $v_i$ is associated with a feature vector $\mathbf{x}_{i} \in \mathbf{X} \in \mathbb{R}^{|V| \times d_F}$ and a label vector $\boldsymbol{y}_c \in \mathbb{R}^{C\times 1}$, where $C$ denotes the number of classes. 
In this work, we focus on node classification task, in which a set of nodes $V_L$ have observed their labels $\mathbf{Y}_{L}$ and the labels of other nodes, denoted as $\mathbf{Y}_{U}$,remaining unobserved. The goal of node classification is to learn a mapping $f: G, \mathbf{X}, \mathbf{Y}_{L} \rightarrow \mathbf{Y}_{U}$ to infer the unobserved labels $\mathbf{Y}_{U}$, utilizing the graph $G$, the node features $\mathbf{X}$ and the observed labels $\mathbf{Y}_{L}$. 

\subsection{Graph Neural Network}

\begin{figure*}[h]
\centering
\includegraphics[width=0.95\linewidth]{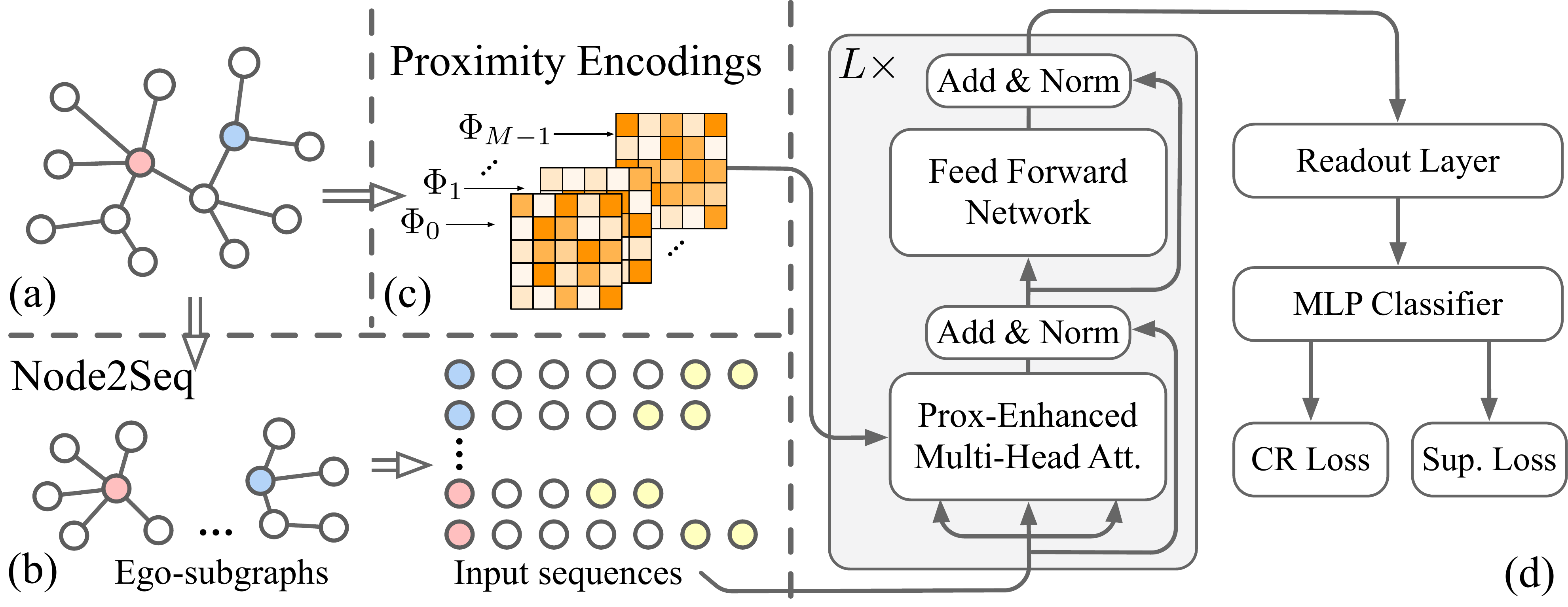}
\caption{Model framework of Gophormer. (a) A sample graph with two example nodes colored red and blue. (b) The Node2Seq process: ego-graphs are sampled from the original graph and converted to sequential data. White nodes are context nodes, yellow nodes are global nodes to store graph-level context. In the example two sample ego-graphs are drawn for each node. (c) The proximity encoding process. (d) The main graph transformer framework of Gophormer.}
\label{fig: Model Framework}
\end{figure*}

GNN model was first proposed by Scarselli et al. to handle both node and graph level tasks \cite{firstGNN}. Then, encouraged by the success of convolutional neural networks (CNNs) \cite{CNN} in the computer vision domain, CNNs was generalized to graph-structured data~\cite{GNN1, GNN2}. Kipf and Welling simplified the spectral GNN model and proposed graph convolutional networks (GCNs) \cite{GCN}. After that, numerous GNN variants have been proposed~\cite{GraphSAGE,ChebNet,GAT,GIN}. Due to its remarkable performance, GNNs have been widely used on many graph applications~\cite{GCN,GIN,PinSAGE,GraphRec,NGCF,GNNSurvey}. 

Despite the great success of existing GNN methods, the message-passing schema of GNNs also suffers from some inherent problems: 
To start with, most GNNs suffer from the over-smoothing problem. It has been observed that deeply stacking the layers often results in significant performance drop for GNNs, such as GCN\cite{GCN} and GAT~\cite{GAT}, even beyond just a few (2–4) layers~\cite{PairNorm}. The reason is that, as the graph convolution operation have been shown as a special form of Laplacian smoothing~\cite{Deeper}, stacking many GNN layers repeats the Laplacian smoothing operation and eventually makes node embeddings indistinguishable. 
Another key issue with existing GNNs is the over-squashing problem~\cite{Bottleneck}. Due to the exponential blow-up in computation path as the model depth increase, it is hard for GNNs to pass information to distant neighbors with a fixed vector size. Hence, GNNs perform poorly if the prediction task is depended on long-range interactions.

\subsection{Graph Transformers}
\label{sec: gt rw}
Transformers~\cite{Transformer} are powerful encoders composed of two major components: a multi-head self-attention (MHA) module and a position-wise feed-forward network (FFN). The MHA module works as follows: Given an input sequence of $\mathbf{H}=\left[\boldsymbol{h}_{1}, \cdots, \boldsymbol{h}_{n}\right]^{\top} \in \mathbb{R}^{n \times d}$ where $d$ is the hidden dimension and $\boldsymbol{h}_{i} \in \mathbb{R}^{d \times 1}$ is the hidden representation at position $i$, the MHA module firstly projects the input $\mathbf{H}$ to query-, key-, value-spaces, denoted as $\mathbf{Q}, \mathbf{K}, \mathbf{V}$, using three matrices $\mathbf{W}_{Q} \in \mathbb{R}^{d \times d_{K}}, \mathbf{W}_{K} \in \mathbb{R}^{d \times d_{K}}$ and $\mathbf{W}_{V} \in \mathbb{R}^{d \times d_{V}}$: 
\begin{equation}
    \mathbf{Q}=\mathbf{H} \mathbf{W}_{Q}, \quad \mathbf{K}=\mathbf{H} \mathbf{W}_{K}, \quad \mathbf{V}=\mathbf{H} \mathbf{W}_{V}.
\end{equation}
Then, in each head $h \in \{1,2,\dots, H\}$, the scaled dot-product attention mechanism is applied to the corresponding $\left \langle \mathbf{Q}_h, \mathbf{K}_h, \mathbf{V}_h\right \rangle$:
\begin{equation}
   \text { head }_{h}=\operatorname{Softmax}\left(\frac{\mathbf{Q}_h \mathbf{K}_h^T}{\sqrt{d_{K}}}\right)  \mathbf{V}_h,
\end{equation}
and the outputs from different heads are further concatenated and transformed to obtain the final output of MHA:
\begin{equation}
    \operatorname{MHA}(\mathbf{H})=\operatorname{Concat}\left(\text { head}_{1}, \ldots, \text { head }_{H}\right) \mathbf{W}_{O},
\end{equation}
where $\mathbf{W}_{O} \in \mathbb{R}^{d \times d}$. In this work, we employ $d_K=d_V=d/H$. 

Due to its remarkable performance, the aforementioned transformer~\cite{Transformer} paradigm has become the go-to architecture in many fields such as natural language processing~\cite{Transformer,BERT,TransformerXL} and computer vision~\cite{VIT,SwinTFM}. In observation of these impressive achievements, a natural idea is to introduce transformer to graphs. Yet, directly applying such feature-based transformers on graphs will lead to the loss of structural information, whereas the structural information is vital in graph mining tasks. Therefore, the key bottleneck of the performance of graph transformer lies in the effective utilization of graph structures. 

To overcome this bottleneck, several structural preserving techniques have been proposed. For example, GT \cite{GT} proposes a generalization of transformer on graph and uses Laplacian eigen-vectors as positional encoding to enhance the node features. SAN \cite{SAN} replaces the static Laplacian eigen-vectors with learnable positional encodings and designs an attention mechanism that distinguishes local connectivity. Graphormer \cite{Graphormer} utilizes centrality encoding to enhance the node feature and uses spatial encoding (SPD-indexed attention bias) along with edge encoding to incorporate structural inductive bias to the attention mechanism. 

However, there are several inherent limitations of existing graph transformers. Above all, the graph is essentially treated as fully-connected with the MHA mechanism calculating attention for all node pairs, causing severe scalability problem. What's more, since the entire graph is treated as only one sequence, this scheme suffers from data scarcity problem if graphs are few (especially for node classification tasks where only one graph is provided in most cases). Therefore, better means of incorporating structural information is required in developing graph transformer models.

\section{The Proposed Method}
\label{sec: method}

In light of the limitations of the full-graph transformers, we propose Gophormer, shown in Figure \ref{fig: Model Framework}, with two simple yet effective designs to incorporate the structural information into traditional transformer. On one hand, Gophormer utilizes the graph topology to generate training inputs via the Node2Seq module, where the sampled ego-graph induced by each center (training) node is converted to a sequential input. Therefore, the local structural information is explicitly preserved in the ego-graphs. On the other hand, Gophormer leverages proximity-enhanced transformer, which utilizes the proximity encoding composed by different views of structural information to calculate attention values. To perform node classification, the 
encoded ego-graph representations are read out and mapped by MLP classifier to the final prediction. The model is optimized by supervised loss and consistency regularization loss. We first introduce how Gophormer prepares sequential data for transformer using the Node2Seq module.

\subsection{Node2Seq}
As stated in Section \ref{sec: gt rw}, current graph transformer methods \cite{GT,Graphormer,SAN} utilize the entire graph to generate input sequences. However, this paradigm is not only memory-consuming but also hard to train, leading to poor performance (further discussed in Section \ref{sec: node2seq exp}). To alleviate such limitations of full-graph-based transformers, we propose to sample ego-graphs as input sequences to capture the local contextual information. Specifically, in each epoch, we sample $S$ ego-graphs for each training node. In this paper, we use GraphSAGE sampling~\cite{GraphSAGE}, which uniformly samples the neighbors of each layer to iteratively construct the ego-graphs with pre-defined maximum depth $D$. Therefore, the local contextual information within $D$-hop is randomly selected as sequential input for the transformer encoder. Note that, users may flexibly design other sampling method to determine the range of contextual nodes. For example, a sampling strategy focusing on high-order neighbors could be a good choice for heterophilic graphs~\cite{H2GNN}.

This strategy is beneficial in several aspects: (1) The ego-graph-based attention can be viewed as full-graph-based attention with random hard attention masks focusing on local information, i.e. nodes outside the sampled ego-graphs are with multiplicative mask value $-\infty$, in this way the structural information is incorporated. (2) Using sampled ego-graphs instead of full-graph can be viewed as a data augmentation technique which not only enjoys performance gain due to transformers' data hungry nature~\cite{datahungry, GPT3} but also alleviates the risk of overfitting. Intuitively speaking, since the sampled ego-graphs from the same node can be different in different epochs, it is harder for the model to fit the training data, enforcing training more generalized model. (3) This strategy apparently enables more scalable training with attention values calculated inside ego-graphs instead of full-graphs. Since the 
input size is reduced from $|V|$ to $|V_{E}|$ (average number of nodes in ego-graphs) with $|V| \gg |V_{E}|$, the memory requirement (quadratic to input size) is greatly reduced.

So far our proposal only focuses on the local structural contexts, i.e. neighbors within pre-defined depth $D$, while the high-order neighbors are neglected. Unfortunately, the importance of such global contextual information has been widely recognized~\cite{HOPE,Bottleneck}. To alleviate this problem, inspired by \cite{BigBird,BERT,Graphormer}, we propose to use global nodes to store global context information. Specifically, we add $n_g$ global nodes to each sampled ego-graph with learnable features $\mathbf{C}= \{ \mathbf{c}_i\in \mathbb{R}^{d\times 1}, i \in 1, \dots ,n_g \}$. These global nodes are shared across all the ego-graphs, which preserves the global context information that assist the node classification task. Hence, in each Gophormer layer, each node not only fuses the information of local neighbors but also the global context information.

In sum, for each epoch, a center node $v_c$ and its induced training ego-graphs $\mathcal{G}_c=\{G^{(s)}_c \mid s\in \{1,2,..., S\}\}$ consist of sampled nodes and global nodes, the input of transformer  $\mathbf{H}_{c,s}^{0}$ is defined as:
\begin{equation}
\mathbf{H}^{0}_{c,s}=\operatorname{Concat}\left(\mathbf{X}^{(s)}_c,\mathbf{C}\right),
\end{equation}
where $\mathbf{X}^{(s)}_c$ denotes the node features of $G^{(s)}_c$. 
\subsection{Proximity-Enhanced Transformer}
\label{sec: pet}
In this section, we introduce the proposed attention mechanism of Gophormer. As introduced in \ref{sec: gt rw}, the attention score of traditional transformers~\cite{Transformer} is calculated by the dot product between encoded query and key embeddings. That is to say, if directly applied, the attention scores between different nodes in each ego-graph are calculated by measuring the similarity between encoded features whereas the vital structural information is ignored. To tackle this, one way to introduce structural inductive bias is to use SPD-indexed attention bias~\cite{Graphormer}. However, this method has two inherent limitations: First, as the name suggests, the SPD only captures the shortest-path relationship, neglecting other structural relationships, e.g. identity~\cite{IDGNN} and higher-order relationships~\cite{HOPE}. Moreover, the SPD-indexed attention bias only reflects the shortest path connectedness between a node pair, neglecting the fine-grained probabilities of connectedness. For example, given two pairs of nodes with SPD as 2, the node pair with more 2-hop path instances apparently have stronger relationships compared to the pair with fewer path instances. Thus, these two pairs of nodes should not be equally treated even if their SPD values are identical. 

We propose proximity-enhanced multi-head attention (PE-MHA) to overcome these limitations. Through depiction of the proposed PE-MHA mechanism, we omit the following notations for brevity: $c$ for center node, $s$ for sample index, and $l$ for transformer layers; since the attention mechanism are identical in each layer and in each ego-graph. Specifically, for a node pair $\left \langle v_i,v_j	\right \rangle$, $M$ views of structural information is encoded as a proximity encoding vector, denoted as $\boldsymbol{\phi}_{ij} \in \mathbb{R}^{M\times 1}$, to enhance the attention mechanism. The proximity-enhanced attention score $\alpha_{i j}$ is defined as:

\begin{equation}
\alpha_{i j}=\frac{\left( \boldsymbol{h}_{i} \mathbf{W}_{Q}\right)\left( \boldsymbol{h}_{j} \mathbf{W}_{K}\right)^{T}}{\sqrt{d}}+ \boldsymbol{\phi}_{ij}^T \boldsymbol{b},
\end{equation}
where $\boldsymbol{b} \in \mathbb{R}^{M\times 1}$ stands for the learnable parameters that calculate the bias of different structural information. The proximity encoding is calculated by $M$ structural encoding functions defined as:
\begin{equation}
    \boldsymbol{\phi}_{ij} = \operatorname{Concat}(\Phi_{m}(v_i,v_j)|m \in 0,1,..,M-1),
\end{equation}
where each structural encoding function $\Phi_m$ encodes a view of structural information. In this paper, we consider two aspects of structural information for each node pair: (1) whether the node pairs are connected at specific order, (2) whether global node exists in this node pair. The proximity encoding functions are defined as:

\begin{equation}
\Phi_m(v_i,v_j)=\left\{\begin{array}{lr}
\mathbf{\tilde{A}}^m[i, j],\quad &\text{if}\quad m< M-1  \\
\mathbb{I}(i,j),\quad &\text{if}\quad m= M-1
\end{array},\right.
\end{equation}
where $\mathbb{I}(i,j)=1$ if global nodes exists in $\left \langle v_i,v_j	\right \rangle$, otherwise 0, $\mathbf{\tilde{A}}=\operatorname{Norm}(\mathbf{A}+\mathbf{I})$ denotes the normalized adjacency matrices with self-loop. In this way, the first $M-1$ dimensions of $\boldsymbol{\phi}_{ij}$ encodes the reachable probabilities from 0-order (identity relationship~\cite{IDGNN}), i.e. whether $v_i$ is $v_j$, to $(M-2)$-order between node $v_i$ and $v_j$. Hence, the fine-grained proximity that reflects different orders of structural information is preserved for each node pair.

\begin{table*}[h]
\centering
\begin{tabular}{c|ccccc} \toprule
Dataset     & \#Nodes & \#Edges & \#Classes & \#Features & Type \\ \midrule
Cora        & 2,708    & 5,278    & 7         & 1,433       & Citation network   \\
Citeseer    & 3,327    & 4,522    & 6         & 3,703       & Citation network    \\
DBLP        & 17,716   & 52,864   & 4         & 1,639       & Citation network    \\
Pubmed      & 19,717   & 44,324   & 3         & 500        & Citation network    \\
Blogcatalog & 5,196    & 171,743  & 6         & 8,189       & Social network    \\
Flickr      & 7,575    & 239,738  & 9         & 12,047      & Social network
\\ \bottomrule
\end{tabular}
\caption{The statistics of the datasets.}
\label{tab: datasets}
\end{table*}
The aforementioned proximity encoding scheme enables the proposed Gophormer with strong expressiveness to handle structural data. In fact, the attention mechanism of Graphormer with SPD-indexed bias can be viewed as a special case of proximity-enhanced attention mechanism with one-hot proximity-encoding of the shortest-path-order. Moreover, as the attention mechanism of Graphormer can be viewed as a special case of PE-MHA, Gophormer also enjoys the merits of representing the aggregation and combination steps in popular GNN models~\cite{Graphormer}.

We follow the GT~\cite{GT} framework to obtain the output of the $l$-th transformer layer, denoted as $\mathbf{H}^{l}$:

\begin{equation}
\begin{array}{lr}
\mathbf{\hat{H}}^{l} =\operatorname{Norm}\left(\operatorname{PE-MHA}\left(\mathbf{H}^{l-1}\right)+\mathbf{H}^{l-1}\right),\\
\mathbf{H}^{l}=\operatorname{Norm}\left(\operatorname{FFN}\left(\mathbf{\hat{H}}^{l}\right)+ \mathbf{\hat{H}}^{l} \right),\end{array}
\end{equation}
where $\text{Norm}(\cdot)$ denotes the layer-norm function. By stacking $L$ layers, Gophormer encodes each node inside the sampled ego-graphs $G^{(s)}_c \in \mathcal{G}_c$ and obtain the embedding of the center node $v_c$, denoted as $\boldsymbol{z}_c^{(s)}$, via a readout function:
\begin{equation}
\boldsymbol{z}^{(s)}_c = \operatorname{Readout}\left(\boldsymbol{h}_{i,c,s}^L \mid v_{i} \in G_c^{(s)} \right).
\end{equation}
The readout function can be implemented by graph pooling functions \cite{GIN,GMT}. Since we are interested in the node property of the center node $v_c$, in this paper, we simply use the center node representation as the final node embedding, i.e. $\boldsymbol{z}^{(s)}_c  = \boldsymbol{h}_{c,c,s}^L$.
\subsection{Optimization}
\label{sec:opt}
 In this section, we introduce how to optimize the Gophormer model. Given the center node representation $\boldsymbol{z}^{(s)}_c$ of a sampled ego-graph $G^{(s)}_c \in \mathcal{G}_c$, Gophormer first adopts a MLP (Multi-Layer Perceptron) function  $f_{m l p}$ with parameters $\Theta$ to predict the node class:
\begin{equation}
\widetilde{\boldsymbol{y}}_c^{(s)}=f_{m l p}\left(\boldsymbol{z}^{(s)}_c, \Theta\right),
\label{eq: cla}
\end{equation}
where $\widetilde{\boldsymbol{y}}_c^{(s)} \in \mathbb{R}^{C \times 1}$ stands for the classification result, $C$ stands for the number of classes. The supervised loss is achieved by the average cross entropy loss of labeled training nodes $V_L$:
\begin{equation}
\mathcal{L}_{\text {sup }}=-\frac{1}{S}  \sum_{v_c\in V_L }\sum_{s=1}^{S}\boldsymbol{y}_c^T \log \widetilde{\boldsymbol{y}}_c^{(s)},
\end{equation}
where $\boldsymbol{y}_c \in \mathbb{R}^{C \times 1} $ is the ground truth label of center node $v_c$. We use consistency regularization \cite{MixMatch,GRAND} to enforce the model make similar predictions for ego-graphs induced by the same center node:
\begin{equation}
\mathcal{L}_{\text {con }}=\frac{1}{S} \sum_{v_c\in V_L }\sum_{s=1}^{S}\left\|\overline{\boldsymbol{y}}_c^\prime-\widetilde{\boldsymbol{y}}_c^{(s)}\right\|_{2}^{2},
\end{equation}
where $\overline{\boldsymbol{y}}_c^\prime$ is the sharpened average distribution of ego-graphs (readers may refer to \cite{GRAND} for more details). The overall loss is obtained by fusing supervised loss $\mathcal{L}_{s u p}$ and consistency regularization loss $\mathcal{L}_{\text {con}}$ with coefficient $\lambda$:

\begin{equation}
\mathcal{L}=\mathcal{L}_{s u p}+\lambda \mathcal{L}_{\text {con}}.
\label{eq: loss}
\end{equation}

\subsection{Inference}
\label{sec: Inference}
Although sampling ego-graphs is an effective data augmentation technique, each sampled ego-graph only preserves partial information within $D$-hop. Therefore, chances are that the important information is lost when sampling testing ego-graphs, leading to inferior results. A straight-forward method to solve this problem is to infer node embeddings based on the full-ego-graph~\cite{GraphSAGE}, i.e. get all the induced nodes within the maximum depth $D$. However, we empirically find that this method leads to inferior results (further discussed in Section \ref{sec: exp_subg_voting}). The potential reason is that the model is trained on ego-graphs while tested on the full-ego-graphs. Therefore, during test time, the model is asked to inference on much larger graphs, leading to performance downgrade~\cite{NonIIDGNN,tang2020investigating}.
\begin{table*}[ht]
\begin{tabular}{c|cccccc} \toprule
Model                              & Cora                & Citeseer            & Blogcatalog         & Pubmed              & DBLP                & Flickr              \\  \midrule
GCN                                & 87.33±0.38          & 79.43±0.26          & 78.81±0.29          & 84.86±0.19          & 83.62±0.13          & 61.49±0.61          \\
GAT                                & 86.29±0.53          & \underline{80.13±0.62}          & 73.20±1.46          & 84.40±0.05          & 84.19±0.19          & 54.29±2.56          \\
GraphSAGE                          & 86.90±0.84          & 79.23±0.53          & 76.73±0.39          & 86.19±0.18          & \underline{84.73±0.28}          & 60.37±0.27          \\
APPNP              & 87.15±0.43          & 79.33±0.35          & \underline{95.63±0.23}          & 87.04±0.17          & 84.40±0.17          & \textbf{93.25±0.24}          \\
JKNet                              & \underline{87.70±0.65}         & 78.43±0.31          & 78.46±1.74          & \underline{87.64±0.26}          & 84.57±0.28          & 53.66±0.40          \\ \midrule
GT-full	    &63.40±0.94	    &58.75±1.06	    &65.32±0.46	    &77.29±0.50	    &78.15±0.41	    &60.77±0.82 \\
GT-sparse	&71.84±0.62	    &67.38±0.76	    &70.65±0.47	    &82.11±0.39	    &81.04±0.27	    &68.59±0.64 \\
SAN         & 74.02±1.01          & 70.64±0.97          & 74.98±0.52          & 86.22±0.43          & 83.11±0.32          & 70.26±0.73          \\
Graphormer & 72.85±0.76          & 66.21±0.83          & 71.84±0.33          & 82.76±0.24          & 80.93±0.39          & 66.16±0.24        \\ \midrule
Gophormer                          & \textbf{87.85±0.10} & \textbf{80.23±0.09} & \textbf{96.03±0.28} & \textbf{89.40±0.14} & \textbf{85.20±0.20} & \underline{91.51±0.28} \\ \bottomrule
\end{tabular}
\caption{Node classification performance (mean±std\%, the best results are bolded and the runner-ups are underlined).}
\label{tab:result-1}
\end{table*}

To bridge the gap between training and testing graphs while alleviating information loss at the same time, we propose a simple yet effective inference method dubbed multi-sample inference: During testing time, the model samples $S^\prime$ ego-graphs for each nodes with the same sampling strategy in training time. The final prediction of a center node $v_c$, denoted as $\widetilde{\boldsymbol{y}}_c$, is obtained by the average of the predictions of  $S^\prime$ sampled ego-graphs:
\begin{equation}
\widetilde{\boldsymbol{y}}_c =\frac{1}{S^\prime} \sum_{s=1}^{S^\prime} \widetilde{\boldsymbol{y}}_c^{(s)}.
\end{equation}
Through this way, the ego-graph sampling method is consistent during training and testing and the loss of information can be alleviated by using larger $S^\prime$. 
\section{Experiments}
\label{sec: Exp}

In this section, we conduct extensive experiments to comprehensively evaluate the proposed Gophormer model.
Following previous works~\cite{GCN, GAT, GraphSAGE}, node classification is selected as the downstream task to evaluate the effectiveness of our proposal. 
After that, ablation studies are conducted to demonstrate the importance of different components. 
Finally, the sensitivity of the model performance on several core parameters are discussed.  

\subsection{Experimental Setup}
\subsubsection{Datasets}

To comprehensively evaluate the effectiveness of Gophormer, we conduct experiments on the six benchmark datasets including four citation network datasets (i.e. Cora, Citeseer, Pubmed~\cite{GCN} and DBLP~\cite{GRACE}), and two social network datasets (i.e. Blogcatalog and Flickr~\cite{BlogFlickr}). We set the  train-validation-test split as 60\%/20\%/20\%. The statistics of datasets are shown in Table \ref{tab: datasets}. 

\subsubsection{Baselines}
To evaluate the effectiveness of Gophormer in graph mining area, we compare it with 8 baseline methods, including five popular GNN methods, i.e. GCN~\cite{GCN}, GAT~\cite{GAT}, GraphSAGE~\cite{GraphSAGE}, JKNet~\cite{JKNet}, and APPNP~\cite{PPNP}, along with four state-of-the-art graph transformers, i.e. GT-Sparse~\cite{GT}, GT-Full~\cite{GT}, Graphormer~\cite{Graphormer}, and SAN~\cite{SAN}. 

\subsubsection{Implementation Details}

We implement Gophormer with Python (3.8.5), Pytorch (1.9.1) and DGL (0.7.1). The code will be made public upon paper publication. We use a 1-Layer-MLP as the $f_{mlp}$ in Eq.\ref{eq: cla} to predict the classes. The global nodes are added to the input of first Gophormer layer's FFN layer
since the feature dimension $d_F$ might unequal with the hidden dimension $d$. We use Adam as optimizer and adopt the ReduceLROnPlateau scheduler of Pytorch. Specifically, the learning rate first goes through a warm-up process, where the learning rate increase linearly from zero to the peak learning rate and then decay with the decay factor according to the validation performance until it reaches the end learning rate. 
\subsubsection{Parameter Settings}
We fix some hyper-parameters for the convenience of tuning work: the dropout is set to 0.5, the weight decay is set to 1e-5, the local contextual information of maximum depth $D$ is set as 2, the hidden dimension $d$ is set to 64 and the number of attention head $H$ is set as 8, the end learning rate is set to 1e-9. Other important hyper-parameters are tuned on each dataset by grid search. The search space of 
learning rate, batch size, number of layers, number global nodes, the number of structural views $M$ for proximity encoding, and the regularization balancing coefficient $\lambda$ in Eq.\ref{eq: loss}, are $\{0.0001, 0.0002\}$, $\{32, 64, 128\}$, $\{1, 2, 3, 4, 5\}$, $\{0, 1, 2, 3, 4\}$, $\{2, 3, 4\}$, $\{0.5, 0.75, 1\}$, respectively. 


\subsection{Node Classification Performance}

The node classification performance of all models are shown in Table \ref{tab:result-1}, from which we have the following observations: (1) Graph transformer baselines are generally outperformed by GNNs especially on the small scale datasets (i.e. Cora and Citeseer). This phenomenon is probably caused by two reasons: First, transformers have much more parameters compared with GNNs and thus they are harder to train especially on small datasets. Second, in most of these datasets, local information is quite important while the attention over the entire graph essentially introduces more noise to the model. 
(2) Gophormer achieves state-of-the-art results and outperforms all graph transformer baselines on all six datasets, proving the effectiveness of our proposed model. Notably, Gophormer achieves better performance than the GNN-based baselines on the small scale datasets (i.e. Cora and Citeseer), indicating the good generalization ability of our proposed model. 
(3) Performance gains of Gophormer on graphs with more nodes or edges (e.g., Flickr, Pubmed and DBLP) are more significant than the ones on the small graphs (e.g., Cora and Citeseer). This is probably due to the data-hungry \cite{datahungry, GPT3} nature of transformers. Since more nodes and edges lead to more data and more possible data augmentation samples, the superiority of Gophormer is more obvious in these datasets.

\subsection{Effectiveness of Node2Seq}
\label{sec: node2seq exp}
\begin{figure*}[h]
\centering
\includegraphics[width=0.95\linewidth]{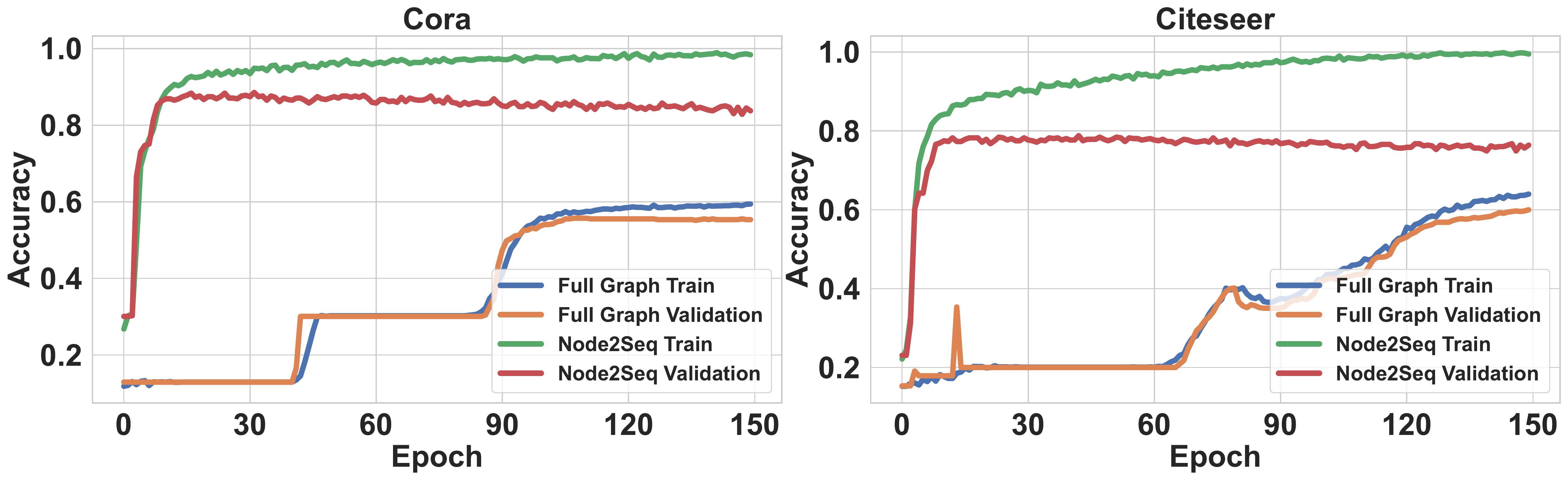}
\caption{The performances of full-graph and Node2Seq data augmentation through training process.}
\label{fig:exp_node2seq}
\end{figure*}

In this section, we investigate the effectiveness of the proposed graph transformer training paradigm Node2Seq by comparing it with the traditional full-graph training paradigm~\cite{GT,SAN,Graphormer}. We first take a closer look at the training and  validation performance through training. As shown in Figure \ref{fig:exp_node2seq}, it is apparently more intractable to train the full-graph based model according to the observed training trajectories. This is reasonable as full-graph training has the penitential to introduce noise from long-distance neighbors, which may disrupt and distract the training process. Meanwhile, compared to the full-graph training, the trajectories of Node2Seq are more smooth and Node2Seq consistently outperforms the baseline from the early beginning, which demonstrates the superiority of the proposed ego-graph training mechanism.

\begin{figure}[h]
\centering
\includegraphics[width=0.95\linewidth]{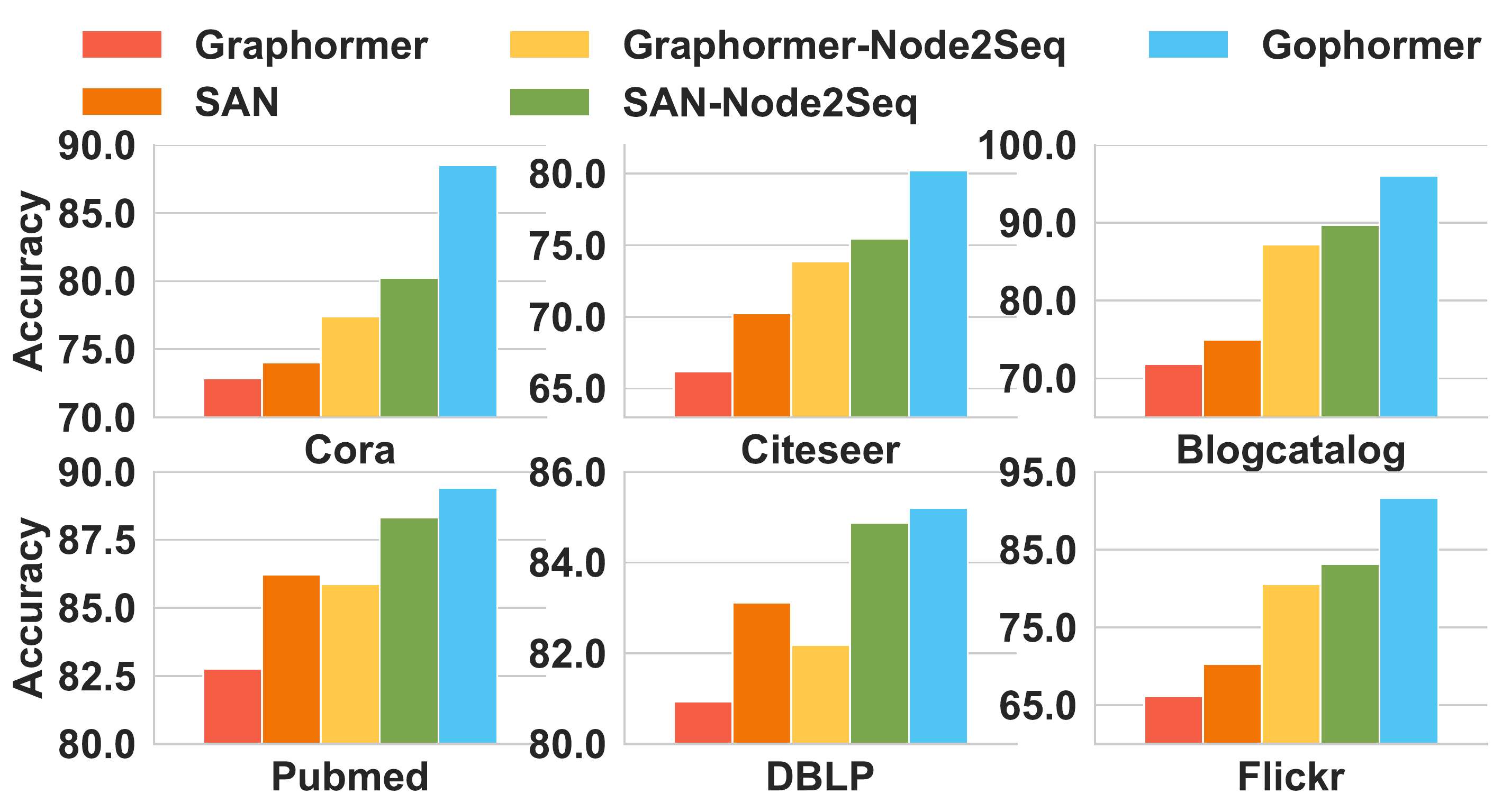}
\caption{The performance evaluation of graph transformers and their variants with Node2Seq training paradigm.}
\label{fig:ego_baselines}
\end{figure}

To further investigate the generality of the Node2Seq technique, we study whether the proposed data augmentation strategy could improve the performance of the two graph transformer baselines. We integrate the proposed Node2Seq into the SAN and Graphormer models denoted as SAN-Node2Seq and Graphormer-Node2Seq. 
Figure \ref{fig:ego_baselines} presents the classification results of different models. 
By enjoying the merits of Node2Seq, SAN-Node2Seq and Graphormer-Node2Seq consistently outperform their vanilla versions by a large margin. Specifically, SAN-Node2Seq outperforms its vanilla version by 6.21\%, 4.81\%, 14.77\%, 2.10\%, 1.77\%, and 12.86\% on the six datasets, respectively. Graphormer-Node2Seq achieves 4.56\%, 7.65\%, 15.37\%, 3.11\%, 1.25\%, 14.38\% gains on the six datasets, respectively. Such a huge performance gain further demonstrates the effectiveness and generality of the proposed sampled ego-graph training paradigm.  
Furthermore, the proposed Gophormer model still consistently outperforms graph transformer baselines with Node2Seq techniques, which proves the effectiveness of other designs within the Gophormer model, e.g. proximity-enhanced transformer, CR loss, and multi-sample inference.    

\subsection{Ablation Studies}
\begin{figure}[h]
\centering
\includegraphics[width=0.95\linewidth]{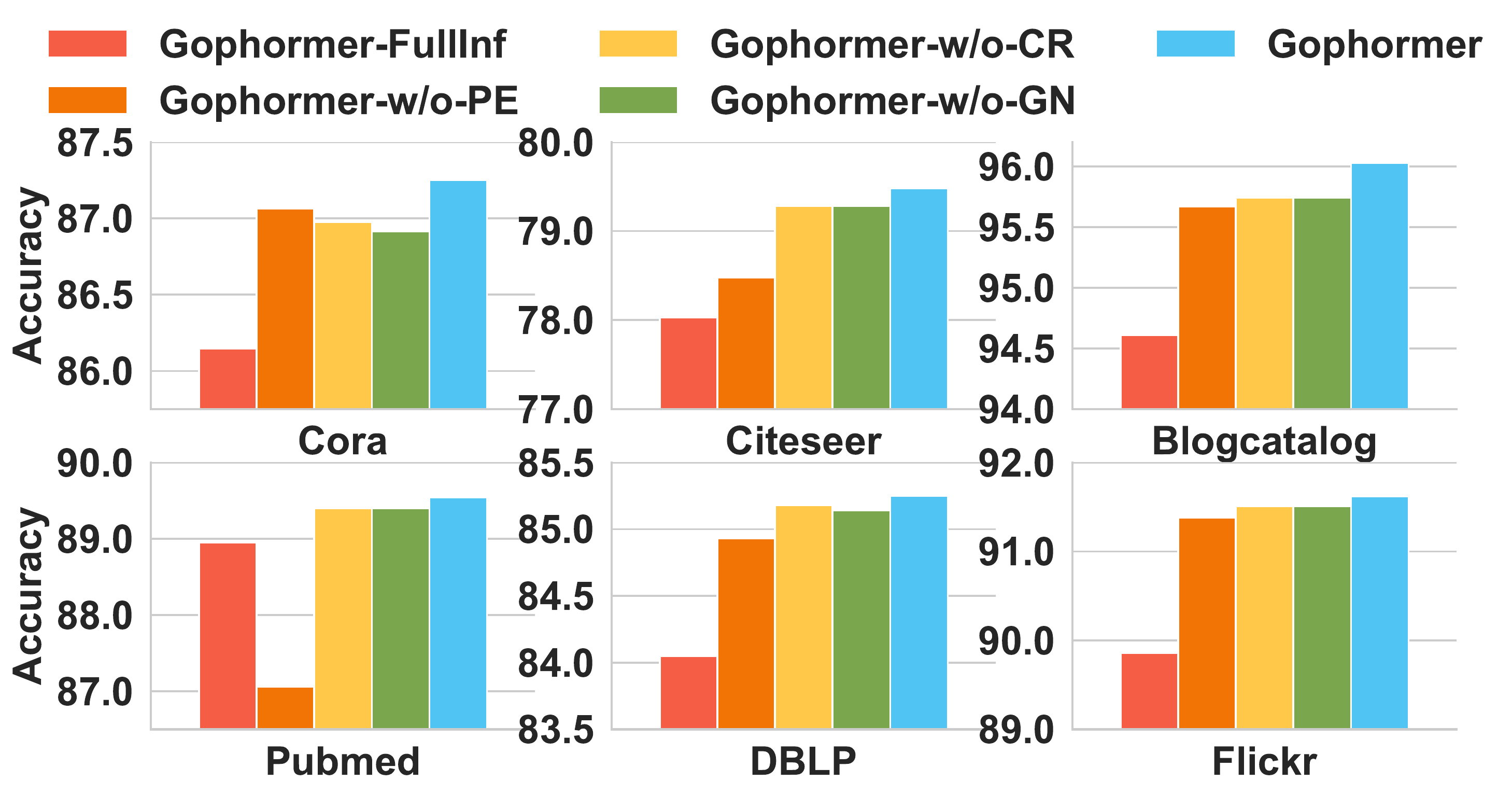}
\caption{The performance evaluation of variants of Gophormer.}
\label{fig:ablation_study}
\end{figure}

In order to verify the effectiveness of different modules in Gophormer, we design five sets of experiments to compare the node classification performance of Gophormer and its variants. The results are shown in Figure \ref{fig:ablation_study}.
\\
\subsubsection{Proximity Encoding Attention Bias}
The vanilla transformer calculates the attention scores purely relying on the node attributes, in which the structural information is ignored in the construction of fully-connected graphs. 
In light of this, the proposed Gophormer model incorporates a proximity-based attention mechanism to better capture the structural bias by considering both feature and structural proximity to learn desirable node representations. To verify the effectiveness of the proximity-based attention mechanism, we design an ablation model Gophormer-w/o-PE without the proximity-based attentions. 
From Figure \ref{fig:ablation_study}, we can observe that Gophormer consistently outperforms Gophormer-w/o-PE on all the six datasets, demonstrating the importance for graph transformer to properly incorporate the structure information.

\subsubsection{Consistency Regularization}

As discussed in Section \ref{sec:opt}, in order to alleviate the uncertainty of ego-graph sampling, Gophormer adopts the consistency regularization loss to enforce the classification distribution learned from different  ego-graphs sampled from the same node to be similar. To investigate the effectiveness of this design, we propose a variant of Gophormer without the ego-graph consistency regularization strategy, namely Gophormer-w/o-CR, and compare its node classification performance with Gophormer. 
From Figure \ref{fig:ablation_study}, one can clearly see that model performance drops after removing the consistency regularization. Without the consistency regularization, the sampled ego-graphs of an identical node could be distinct, leading to the different predictions and inferior classification performance. Our proposal can alleviate the uncertainty of sampling ego-graphs and thus boost the performance of Gophormer.

\subsubsection{Global Nodes}
As discussed in Section \ref{sec: pet}, Gophormer utilizes a set of shared global nodes across different ego-graphs to convey global contextual information. 
To evaluate the importance of these global nodes, we design another ablation model Gophormer-w/o-GN without the global nodes. 
Figure \ref{fig:ablation_study} shows the classification performance of this variation model. 
After removing the global nodes, the performance consistently exhibits a significant decline over all datasets. 
The ego-graphs are generated within a fixed hops from the center nodes and then fed into graph transformer. 
Without adding global nodes, the receptive field of model is limited to a localized small area and further degrades the model preference.  
Global nodes can be viewed as the intermediaries to exchange global information across different ego-graphs, which are capable of providing external global context as complementary to the local information preserved in the ego-graphs. 
By simultaneously enjoying the merits of local and global information, Gophormer achieves desirable performance over all datasets.  

\subsubsection{Multi-sample Inference}
\label{sec: exp_subg_voting}

As discussed in section \ref{sec: Inference}, we propose a multi-sample inference strategy to handle uncertainty and relieve the structural inconsistency between the training and testing sets. 
To verify these claims, another ablation model Gophormer-FullInf without the multi-sample strategy is designed.   
Gophormer-FullInf is trained on sampled ego-graphs and tested on the full-ego-graph without sampling. 
From the results shown in Figure \ref{fig:ablation_study}, we empirically find that the multi-sample schema contributes to boosting the classification performance compared to the Gophormer-FullInf model. 
The results reveal that the consistent graph sampling method between training and testing sets is crucial to achieve desirable performance, and the proposed multi-sample inference strategy is capable of achieving promising results.

\subsubsection{Evaluation on Inference Strategies}
\begin{figure}[ht]
\centering
\includegraphics[width=0.95\linewidth]{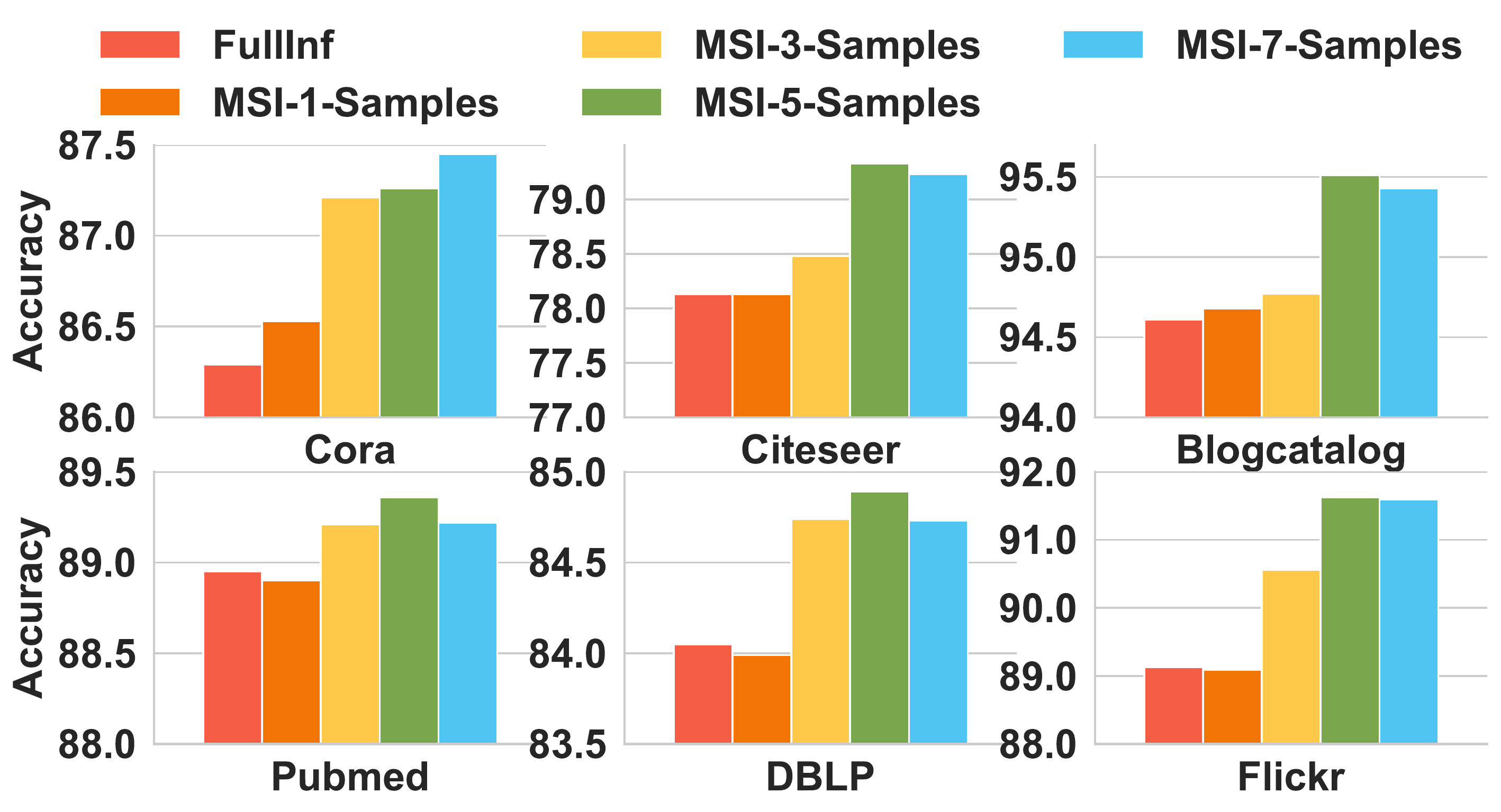}
\caption{The impact of varying number of samples used multi-sample inference settings.}
\label{fig:pa_ValInf}
\vspace{-2mm}
\end{figure}

We demonstrated the effectiveness of the proposed multi-sample inference schema in section \ref{sec: exp_subg_voting}. 
In this section, we aim to further reveal the potential reasons why this strategy will work. 
Two types of ablation models are proposed, including the FullInf which infers the predictions based on the full-ego-graphs, and MSI-$k$ denotes the proposed multi-sample inference strategy with $k$ samples. 
From the results shown in Figure \ref{fig:pa_ValInf}, we can see that FullInf and MSI-1-sample settings achieve the worst performance. 
This is reasonable as FullInf cannot handle the structural inconsistency between graph samples in training set and testing set, while MSI-1-sample makes decisions solely based on a single sample and thus its performance is severely hindered by the uncertainty. 
Other three MSI variations with larger number of samples achieve better performance. 
With the increase of $k$, the performance first increases and then keeps steady. 
It means that the incorporation of more sampled ego-graphs in the inference phase contributes to better classification performance at the beginning. 
When the real category distribution is fully revealed, more samples cannot further improve model performance.  
\subsection{Parameter Analysis}
In this section, we study the performance sensitivity of the proposed Gophormer model on two core hyper-parameters: the number of global nodes $n_{g}$ and the number of transformer layers $L$. 

\subsubsection{Global Nodes}
\begin{figure}[ht]
\centering
\includegraphics[width=1.\linewidth]{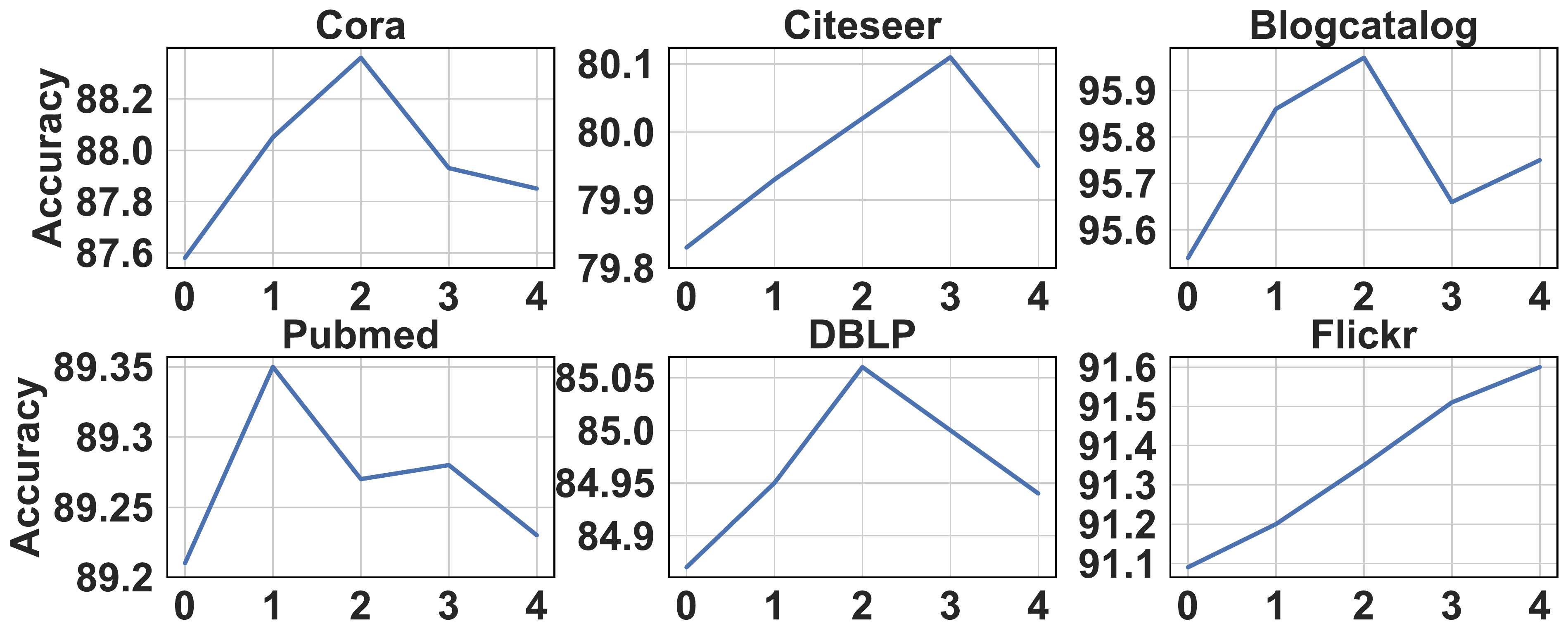}
\caption{The classification performance of Gophormer with different global nodes number $n_g$.}
\label{fig:pa_GN}
\vspace{-2mm}
\end{figure}

The ego-graph contains only the local structural contexts while neglects the high-order information. Therefore, we propose to add $n_{g}$ shared global nodes into the sampled ego-graphs to introduce the global structural information. Here we aim to study the impact of varying the number of global nodes.  Figure \ref{fig:pa_GN} presents the node classification performance of Gophormer with $n_{g}$ varying from 0 to 4. From the results, one can see that with the increase of $n_{g}$, the performance increases until reaches at a peak and then decreases. This is reasonable as suitable number of global nodes are capable of incorporating proper global information, while excessive global nodes may introduce too many parameters which slow down the training speed and hurt the generalization ability. Thus, the number of global nodes should be carefully decided to achieve the optimal performance. 

\subsubsection{Number of Gophormer Layers}
\begin{figure}[h]
\centering
\includegraphics[width=1.\linewidth]{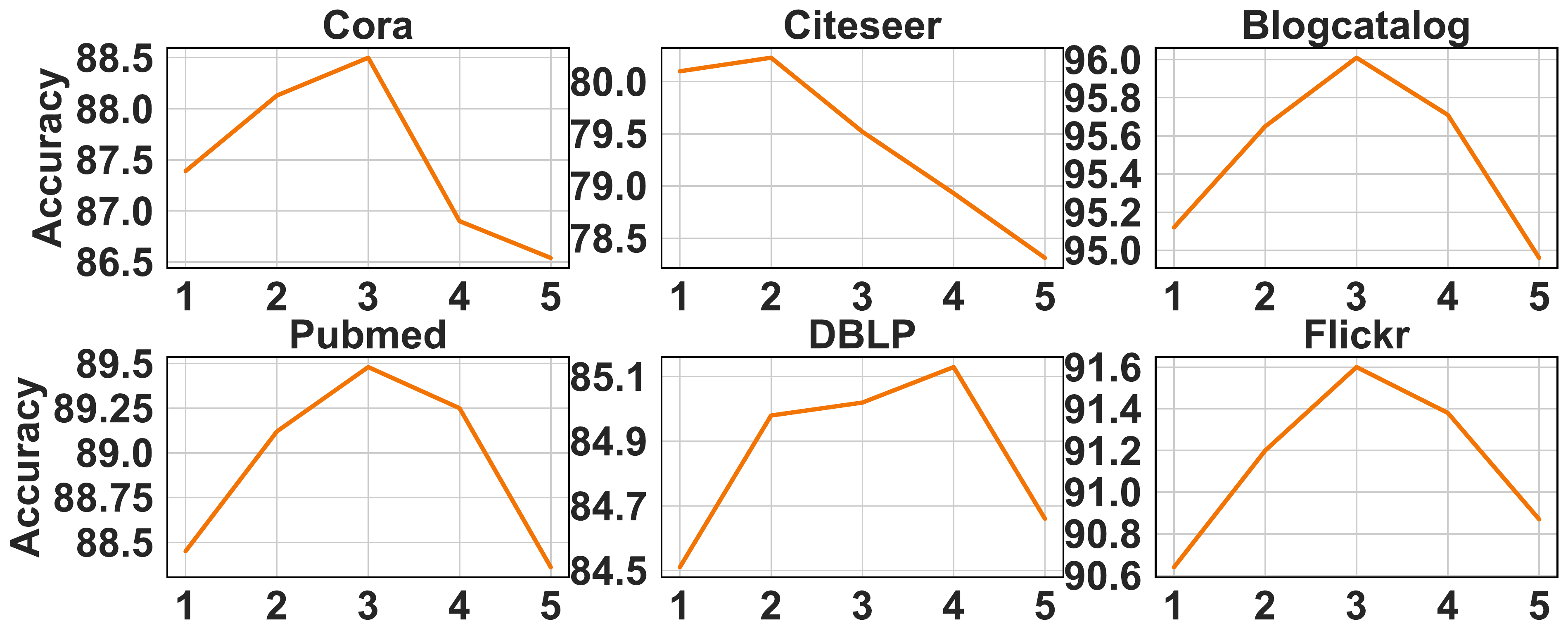}
\caption{The classification performance of Gophormer with different numbers of network layer $L$.}
\label{fig:pa_NL}
\end{figure} 


We further study the influence of the number of Gophormer layers $L$ on the classification performance. We vary $L$ from 1 to 5 and exhibit the results in Figure \ref{fig:pa_NL}. It is obvious that with the increase of $L$, the performance significantly increases at the beginning, since stacking Gophormer layers enlarges representation learning capacity which is beneficial in capturing sophisticated structural patterns. However, deeper models also suffer from serious challenge of over-fitting, which is responsible for the performance decline with 5 transformer layers over all the datasets. 
Gophormer on small datasets, like Cora and Citeseer, achieves best performance with $L$ set to 2 or 3. Meanwhile on other larger datasets, the best performance is often achieved when $L$ is set to 3 or 4. Hence, the number of transformer layers should be carefully chosen based on the graph scale and characteristics. 

\section{Conclusion}
\label{sec: Conc}
The transformer architecture has shown dominant performance on CV and NLP tasks, but not yet in the graph field especially on node classification tasks. In this work, we propose a novel ego-graph-based transformer model dubbed Gophormer, which effectively incorporates the structural information by Node2Seq module and the proximity-enhanced attention mechanism. 
We also design consistency regularization loss and multi-sample inference to alleviate the negative impacts of sampling. Extensive experiments are conducted to show the effectiveness of the proposed Gophormer model, revealing the promising future of the emerging field of graph transformers.

\newpage
\bibliographystyle{ACM-Reference-Format}
\bibliography{Gophormer_bib.bib}

\end{document}